\documentclass[lettersize,journal]{IEEEtran}
\usepackage{amsmath,amsfonts}
\usepackage{algorithm}
\usepackage{algorithmicx}
\usepackage{algpseudocode}
\usepackage{xcolor}
\usepackage{array}
\usepackage[caption=false,font=normalsize,labelfont=sf,textfont=sf]{subfig}
\usepackage{textcomp}
\usepackage{stfloats}
\usepackage{url}
\usepackage{verbatim}
\usepackage{graphicx}
\usepackage{multirow}
\usepackage{threeparttable}
\def\BibTeX{{\rm B\kern-.05em{\sc i\kern-.025em b}\kern-.08em
    T\kern-.1667em\lower.7ex\hbox{E}\kern-.125emX}}
\usepackage{balance}
\usepackage{hyperref}
\usepackage[numbers,sort&compress]{natbib}
\newcommand{\tabincell}[2]{\begin{tabular}{@{}#1@{}}#2\end{tabular}}

\usepackage{setspace} 

\makeatletter
\def\thanks#1{\protected@xdef\@thanks{\@thanks
        \protect\footnotetext{#1}}}
\makeatother

\begin{document}

\title{DynPL-SVO: A Robust Stereo Visual Odometry for Dynamic Scenes}


\author{Baosheng Zhang, Xiaoguang Ma, Hong-Jun Ma and Chunbo Luo}

\thanks{Baosheng Zhang and Xiaoguang Ma are with the College of Information Science and Engineering, Northeastern University, 110819 Shenyang, China (Xiaoguang Ma is the corresponding author, e-mail: maxg@mail.neu.edu.cn).}

\thanks{Hongjun Ma is with the School of Automation Science and Engineering, South China University of Technology, Guangzhou 510641, China (e-mail: mahongjun@scut.edu.cn).}

\thanks{Chunbo Luo is with the School of Information and School of Information and Communication Engineering, University of Electronic Science and Technology of China, Chengdu 611731, China (e-mail: c.luo@uestc.edu.cn).}

\maketitle
\begin{abstract}
    Most feature-based stereo visual odometry (SVO) approaches estimate the motion of mobile robots by matching and tracking point features along a sequence of stereo images. However, in dynamic scenes mainly comprising moving pedestrians, vehicles, etc., there are insufficient robust static point features to enable accurate motion estimation, causing failures when reconstructing robotic motion.
In this paper, we proposed DynPL-SVO, a complete dynamic SVO method that integrated united cost functions containing information between matched point features and re-projection errors perpendicular and parallel to the direction of the line features. Additionally, we introduced a \textit{dynamic} \textit{grid} algorithm to enhance its performance in dynamic scenes. The stereo camera motion was estimated through Levenberg-Marquard minimization of the re-projection errors of both point and line features. Comprehensive experimental results on KITTI and EuRoC MAV datasets showed that accuracy of the DynPL-SVO was improved by over 20\% on average compared to other state-of-the-art SVO systems, especially in dynamic scenes.

\end{abstract}


\section{Introduction}
\IEEEPARstart{V}{isual} odometry (VO) is a popular research topic in the fields of robotics, autonomous driving, and augmented reality. It uses various types of cameras to estimate its mobile motion and reconstruct surrounding map \cite{2023sgslam} \cite{pumarola2017pl}. The stereo visual odometry (SVO) has attracted more attention recently due to its low cost, robustness, and wide applicability for both indoor and outdoor scenes \cite{2022univio} \cite{2020obser}.

Most VO systems rely solely on point features for motion estimation, as they are easy to detect, track, and handle \cite{mur2017orb} \cite{kitt2010visual}. However, point features have poor anti-interference ability and are sensitive to lighting variations, occlusion, and rapid motion, limiting the performance of point-feature-based SVO. Introducing line features can effectively address these issue by providing more constraints, improving the accuracy and stability of camera pose estimation. Moreover, in dynamic scenes, it is difficult to extract sufficient static point features to estimate pose of robots, and line features extracted from static regions in the scene can complement the deficiencies of point features in dynamic scenes. 
\begin{figure}[!t]
    \centering
    \includegraphics[width=3.5in]{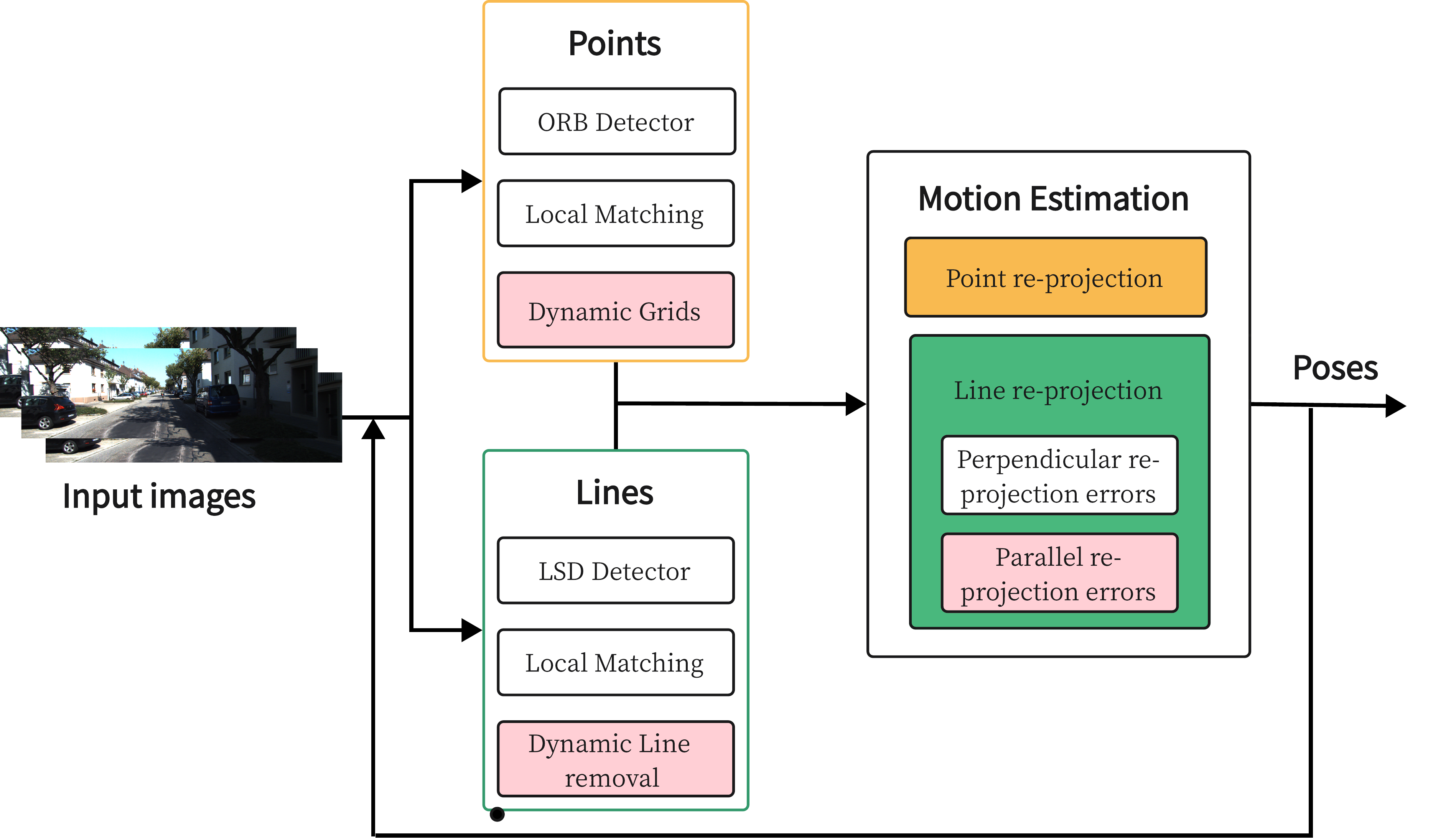}
        \caption{Overview of the DynPL-SVO.}
    \label{fig:intro}
\end{figure}

Dealing with dynamic scenes is a significant challenge for traditional VO methods, as they often fail to achieve accurate inter-frame matching, causing low motion estimation accuracy. This was commonly solved by accurately removing dynamic outliers in images before pose optimization, meaning that the system must remove features introduced by the dynamic objects and rely solely on trusted static features for the motion estimation. Therefore, accurately extracting and removing the dynamic features is critical for improving  the performance of VO systems in the dynamic scenes.

The robust constraints and random sample consensus (RANSAC) had been applied to remove outliers (dynamic points) \cite{2017Hierarchical}, and Mask R-CNN is an excellent semantic segmentation network in detecting dynamic objects in scenes \cite{2017Mask}. However, the above methods have serious efficiency shortcomings, limiting their application in computer vision problems that require real-time performance. As an online camera pose estimation system, VO estimates the camera's current pose based on the previous frame motion models, and geometric constraints can be used to identify dynamic regions in the scene. In this paper, we proposed \textit{dynamic grid} approach which used spatial geometric constraints as a prerequisite for mitigating the impact of dynamic scenes on VO, and did not require additional depth information or an explicit understanding of the scene.

In this paper, we proposed DynPL-SVO, a robust and complete SVO system to fully use the structural and geometric information of images to improve accuracy, especially for dynamic scenes.

The main contributions of this paper were listed below:
\begin{itemize}
    \item We proposed a complete SVO system, named DynPL-SVO, which utilized both point and line features and was capable of effectively coping with dynamic scenes using only stereo RGB images from stereo cameras. 
    \item The re-projection errors of the point features and re-projection errors perpendicular and parallel to line features were jointly used to construct a unified cost function for pose optimization, resulting in superior robustness compared to conventional feature-based approaches.
    \item A \textit{dynamic grid} approach was carefully designed to identify dynamic regions and remove dynamic features, and improvement of 13.6\% and 2.3\% on absolute pose error (APE) and relative pose error (RPE), respectively, were made on highly dynamic scenes, such as KITTI-01, 05, and 09.
    \item Extensive comparative experiments were conducted on both the KITTI and the EuRoC MAV dataset to validate performance of the DynPL-SVO. The results showed that, the translation RMSE drifts of the DynPL-SVO were improved by 13.8\%, 30.0\%, and 24.8\% as compared to those of PL\_SLAM front end, ORB\_SLAM2 front end, and ORB\_Line SLAM front end, respectively.
\end{itemize}

\section{Related Work}
VO methods could be divided into direct-based \cite{engel2014lsd} \cite{engel2017direct} and feature-based \cite{mur2017orb} according to how visual measurements were processed. Direct-based methods used intensity of each pixel to compute camera's motion by minimizing photometric errors, without detecting and matching specific features. However, most VO systems used feature-based methods due to their high robustness and estimation accuracy, wherein feature descriptors were used to detect and track point features and estimate motion by minimizing the re-projection errors between detected features and their corresponding projected features from frames. To ensure real-time performance and reliability of the VO system, many researchers used ORB \cite{mur2017orb} \cite{zuo2017robust}, which provides robust and accurate motion estimation in rich-texture scenes even with only point features. However, the accuracy of motion estimation with insufficient point features can greatly degrade in poor-texture scenes. As a result, many researchers introduced line features to improve robustness and accuracy \cite{zuo2017robust} \cite{zhang2015building} \cite{gomez2016pl}, and several methods have achieved satisfactory results in detecting straight-line features, such as FLD \cite{lee2014outdoor}, EDLine \cite{akinlar2011edlines}, and LSD \cite{von2008lsd}.

One common and concise way to represent line features is to use two endpoints to model a 3D line \cite{koletschka2014mevo}. To avoid optimization constraints brought by the over-parameterized representation of line features, researchers \cite{zuo2017robust} \cite{he2018pl} applied orthogonal representation \cite{bartoli2005structure} and $Pl\ddot{u}cker$ coordinate \cite{gomez2016pl} to transform and optimize line features, respectively.

Optimization of line features required various representations with corresponding cost functions. Koletschka \cite{koletschka2014mevo} used the Euclidean distance sum of equally spaced sampling points on the line segments as cost functions for the line features. In most methods \cite{he2018pl, zuo2017robust, gomez2016pl, gomez2016robust}, the line re-projection errors were defined as the Euclidean distance from the endpoints of the detected line features in a current frame to their projections in a previous frame. All these cost functions could be easily computed using $Pl\ddot{u}cker$ coordinates and the distance formula between points and lines. However, previous studies only considered structural information perpendicular to the direction of line features, and implementing a cost function that integrated re-projection errors perpendicular and parallel to line features could substantially enhance SVO performance.

Estimating motion in dynamic scenes poses a significant challenge to VO systems. Several works \cite{sahdev2018indoor, ouyang2021visual} have addressed this issue by leveraging vision information to fuse other sensors, such as IMU and wheel odometry. Additionally, numerous methods \cite{kim2020moving} used RGB-D information to identify dynamic areas where unstable point features were removed, allowing only stable static point features to be retained during optimization. More recently, deep learning-based solutions were combined with VO systems to provide feasible solutions to handle dynamic scenes \cite{sun2021unsupervised}. However, all these methods impose strict requirements on scenario conditions and computer resources, and a novel SVO approach that could achieve accurate motion estimation in dynamic scenes without relying on image depth information or assistance from other sensors would be highly desirable.

\section{Detection and Pre-Processing}
\subsection{Feature detection and matching}
\subsubsection{Point features}
\begin{figure}[!t]
    \centering
    \includegraphics[width=3.5in]{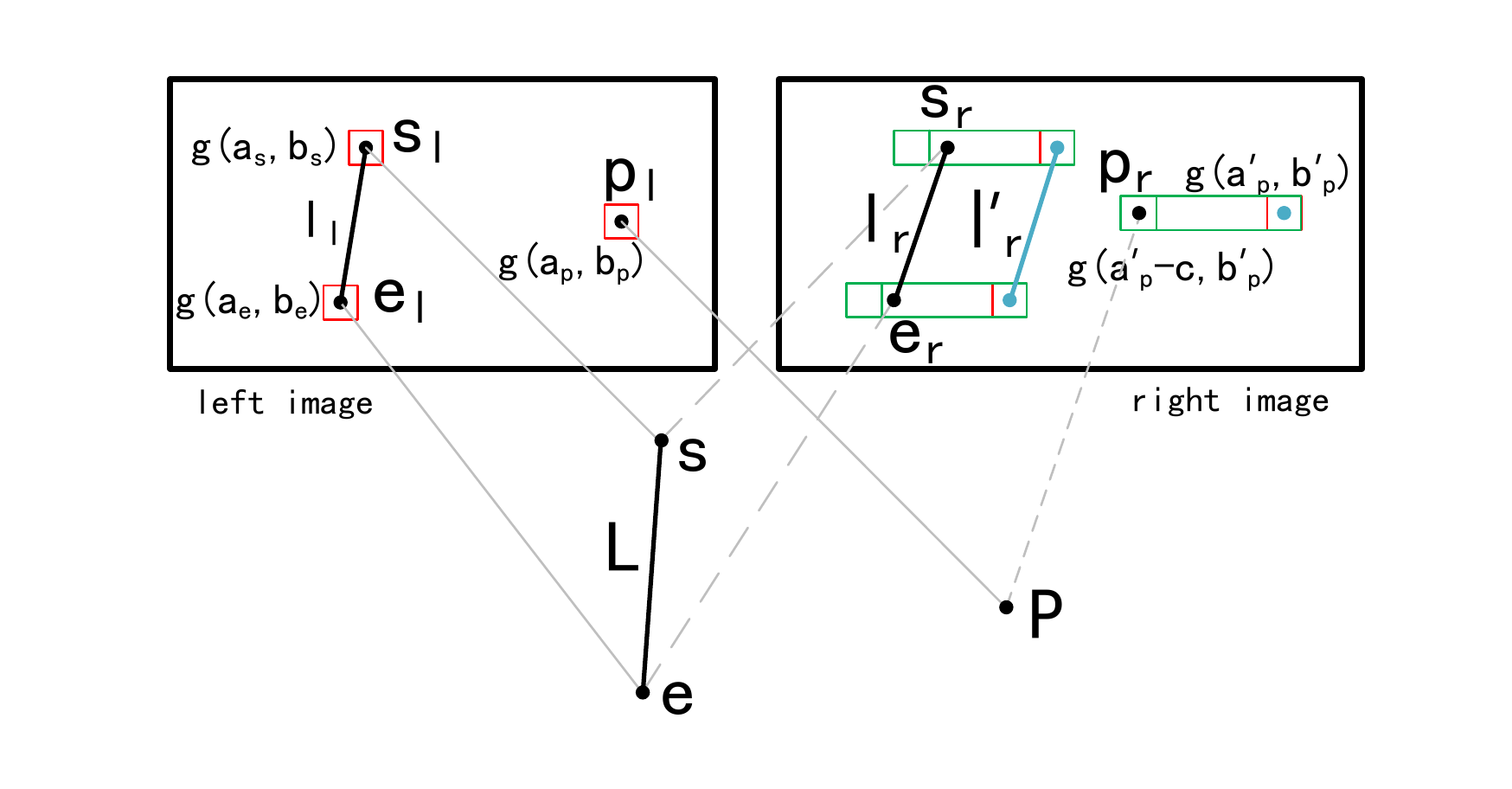}
    \caption{The feature matching process between the left and right images in a stereo frame. We assumed that the point $p_l$ detected in the left image in the $grid(a_p, b_p)$ (red square), followed the epipolar constraint and imaging principle of the stereo camera. The point (such as $p_r$) to be matched in the right image should be located at the range from $grid(a^{\prime}_p-c,  b^{\prime}_p)$ to $grid(a^{\prime}_p, b^{\prime}_p)$ in the right image, i.e., the green rectangular area. Similar to point matching, if the endpoints of the line $l_r$ detected in the right image met the matching rule, it would be a candidate line of $l_l$.}
    \label{fig1:det}  
\end{figure}

After detecting an ORB point feature $p_l(u_l, v_l)$ in the left image and its corresponding feature $p_r(u_r, v_r)$ in the right image of a stereo frame, as illustrated in Figure \ref{fig1:det}, the next step was to match these features between the two images. To achieve this, we evenly divided each image into 64$\times$48 grids and stored the point features according to their grid positions within the images. It is essential to note that the feature matching must follow the epipolar constraint, i.e., $v_l = v_r$, and comply with the imaging principle, i.e., $u_l > u_r$. The point features required for matching in the right image were limited to horizontal grids, ranging from $grid(a^{\prime}_p-c, b^{\prime}_p)$ to $grid(a^{\prime}_p, b^{\prime}_p)$. Additionally, a mutual consistency check was performed, meaning that only matches with corresponding best-left and best-right matches were deemed valid.

\begin{figure}[!t]
    \centering
    \includegraphics[width=3.25in]{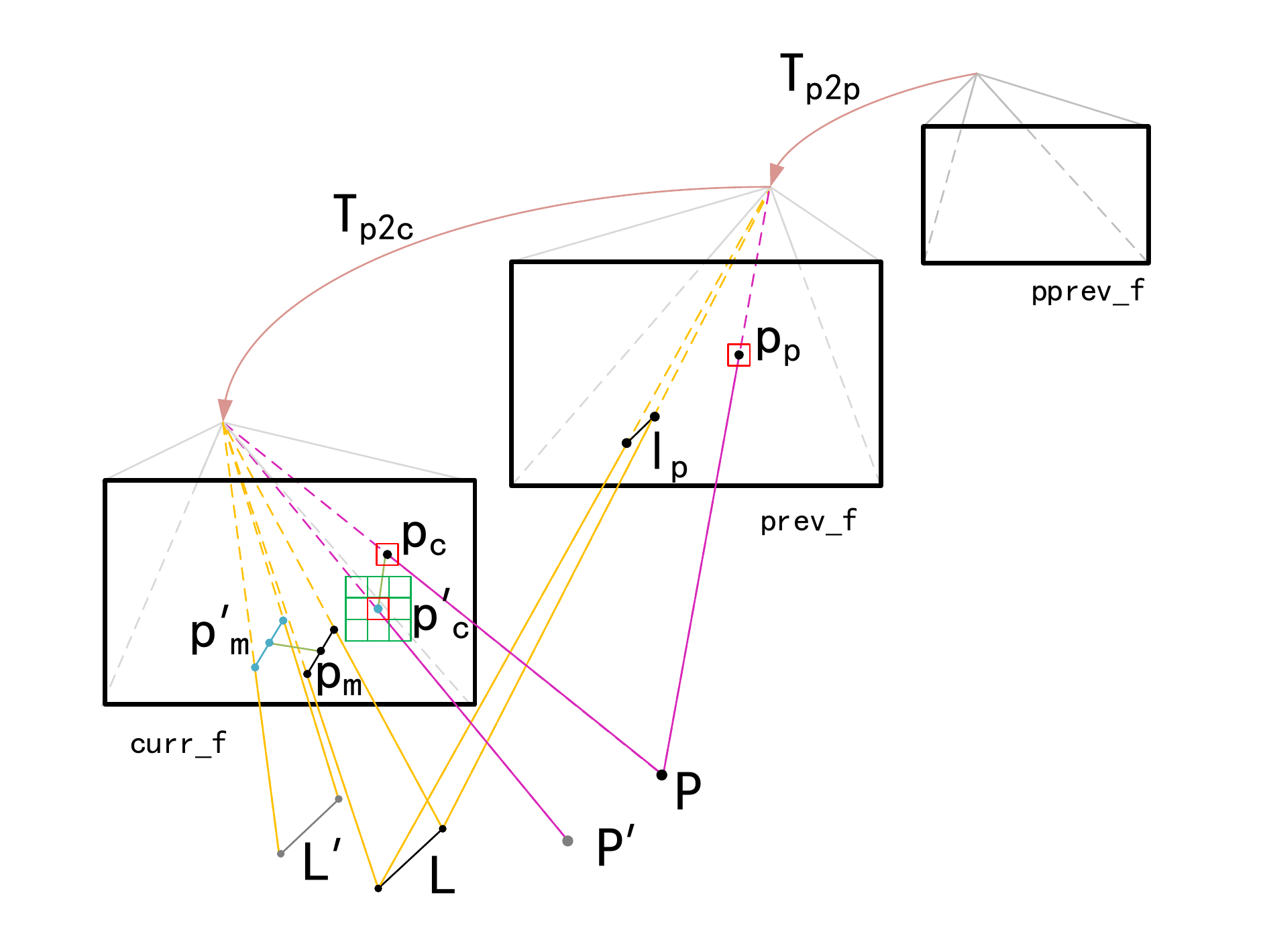}
    \caption{The feature tracking process between adjacent frames. $T_{p2p}$ denoted pose transformation between the previous frame $prev\_f$ and its preceding frame $pprev\_f$. $T_{p2c}$ denoted the pose transformation between $prev\_f$ and the current frame $curr\_f$. The dynamic spatial point $P$ and spatial line $L$ moved to $P'$ and $L'$ during sampling time, respectively. For line features $L$, we defined their re-projection errors in a similar way with point features by utilizing the Euclidean distance between midpoints $p'_m$ and $p_m$ of matched and estimated line features. Based on this, we classify a feature as dynamic or not.}
    \label{fig2:det}
\end{figure}

Figure \ref{fig2:det} illustrated the feature tracking process between adjacent frames. To mitigate the impact of dynamic object features on our estimation results, we utilized a motion model to estimate the locations of the matched point features in the current frame $curr\_f$. Specifically, we employed a uniform motion model represented by a pose transformation matrix $T_{p2p}$ between the previous frame $prev\_f$ and its preceding frame $pprev\_f$ as the initial state of the motion model.
Given that dynamic objects exhibited abnormal motion relative to static scenes, the dynamic spatial point $P$ moved to $P'$ during sampling time, resulting in larger re-projection errors for dynamic features $p'_c$ compared to estimated features $p_c$ predicted by the motion model. 
To identify \textit{dynamic grid}s, we computed and averaged the sum of squared Euclidean distances between matched point features and estimated features in all relevant grids. If this value exceeded a threshold, we classified the grid and its surrounding eight grids in green as \textit{dynamic grid}s, and the point features within these grids were identified as dynamic point features. This enabled us to accurately track and estimate feature locations in complex and dynamic scenes.

Algorithm \ref{det:alg} showed detailed process of proposed dynamic region marking based on \textit{dynamic grid} algorithm.

\begin{algorithm}
    \caption{: Dynamic region marking using the \textit{dynamic grid}.}
    \begin{algorithmic}[1] 
        \Require Uniform motion model $T_{p2p}$ of the previous frame $prev\_f$, the matched point feature set between the current frame $curr\_f$ and $prev\_f$;
        \Ensure The location of the \textit{dynamic grid}.
            \State Divide $curr\_f$ evenly into 64$\times$48 grids and only keep $n(n<=8)$ point features $p_j$ in each grid $g_i$;
            \For {$ each $ $ g_i \in curr\_f$} 
                
            \For {$each $ $ p_j \in g_i $}
                \State $e_{g_i}$ += $PointErr(p_j,T_{p2p}) / n$
            \EndFor

            \If{$e_{g_i} > \rho$}
                \State $GRIDS\_LOCATION $ += $\{(x_{g_i}-1, y_{g_i}-1)\sim (x_{g_i}+1, y_{g_i}+1)\}$
            \EndIf
            \EndFor
        \State \Return{$GRIDS\_LOCATION $}          
    \end{algorithmic}
    \label{det:alg}
\end{algorithm}

\subsubsection{Line features}
In this work, line features were extracted from images using LSD and represented using LBD. Similar to the point feature matching process between the left and right images as shown in Figure \ref{fig1:det}, we grouped the line features in the right image that passed through the same grids. We assumed that the endpoints of the line feature $l_l$ in the left image were located in $(a_s, b_s)$ and $(a_e, b_e)$. In the right image, only lines within corresponding grids were considered as candidate matches. To ensure the accuracy of the matches, we applied a line matching rule and performed a mutual consistency check, as illustrated in Figure \ref{fig1:det}. Only matches that satisfy these requirements were selected for further processing.

In Figure \ref{fig2:det}, we showed the use of re-projection errors between the estimated line midpoint $p_m$ and matched line midpoint $p'_m$ as the criterion for dynamic line feature tracking between adjacent frames. Once the errors exceed a pre-set threshold, the line features were identified as dynamic ones and removed. Furthermore, effective \textit{dynamic grid}s were obtained in the scene, as mentioned above.

\subsection{Representation of the Line Features}
We assumed that the homogeneous coordinates of line endpoints were represented by $\bar{X_s}(x_1, y_1, z_1, w_1)$ and $\bar{X_e}(x_2, y_2, z_2, w_2)$, with their inhomogeneous counterparts denoted as $X_s$ and $X_e$. The $Pl\ddot{u}cker$ coordinates of the line $L$ could then be constructed using the following formula:

\begin{equation}
    L = \begin{bmatrix}
    X_s \times X_e \\ w_2X_s - w_1X_e
    \end{bmatrix} = \begin{bmatrix}
     \mathbf{n} \\ \mathbf{d}
    \end{bmatrix}\in \mathbb{R}^6 
\end{equation}
, where $\mathbf{d}$ represented the direction vectors of the line, and $\mathbf{n}$ denoted the normal vectors of the plane determined by the lines and the origin. Specifically, $\mathbf{n}^T \times \mathbf{d} = 0$. The $Pl\ddot{u}cker$ coordinates for $L$ could also be extracted from the dual $Pl\ddot{u}cker$ matrix $T^*$, which was defined as follows:

\begin{equation}
    T^* = \begin{bmatrix}
    \mathbf{d}^\wedge  & \mathbf{n}\\
    -\mathbf{n}^T & 0
    \end{bmatrix}
\end{equation}
, where $\wedge$ denoted transformation between vectors and antisymmetric matrices.
The representation of the $Pl\ddot{u}cker$ coordinates was chosen due to its convenience for line feature projection, transformation, and Jacobian usage.

\section{Motion Estimation}
\subsection{Problem Statement}\label{4A}
The primary objective of VO systems is to find an optimal transformation that can satisfy the projection constraints for the corresponding features with high accuracy. This can be achieved by solving a non-linear least-squares equation formed by the projection constraints of corresponding features between the adjacent frames.


Compared to other VO systems that rely solely on the re-projection errors of point features and the re-projection errors perpendicular to the direction of line features. In this paper, we proposed DynPL-SVO, a complete dynamic SVO that integrated cost functions containing information between matched point features and re-projection errors perpendicular and parallel to the direction of the line features, effectively leveraging the rich structural information contained in the endpoints of line features detected by LSD, leading to increased robustness and accuracy. The non-linear least-squares equation for the proposed method was shown in following:
\begin{equation}
\begin{split}
\label{equ:est1}
\xi ^* &=\underset{\xi}{arg\ min} \Bigg[ \sum_{i=1}^{m}{e_{i}^{p}(\xi)}^{T}\Sigma _{e_{i}^{p}}^{-1}e_{i}^{p}(\xi) +  \\
&\sum_{j=1}^{n}{e_{j}^{l_{pe}}(\xi)}^{T}\Sigma _{e_{j}^{l_{pe}}}^{-1}e_{j}^{l_{pe}}(\xi) +  \sum_{k=1}^{q}{e_{k}^{l_h}(\xi)}^{T}\Sigma _{e_{k}^{l_{pa}}}^{-1}e_{k}^{l_{pa}}(\xi) \Bigg]  
\end{split}
\end{equation}
, where $m$, $n$, and $q$ respectively denoted the numbers of points, lines, and the numbers of complete line features extracted within the image. The equation included point re-projection errors ($e_i^p$), re-projection errors perpendicular to the line direction ($e_k^{l_{pe}}$), and re-projection errors parallel the direction of line features ($e_j^{l_{pa}}$). The $\Sigma^{-1}$ matrices in (\ref{equ:est1}) represented the inverse covariance matrices related to the uncertainty of each re-projection error term.
    
The re-projection errors of point features were defined as the distance between projected point features from the previous frames and the corresponding ones in the current frames:
\begin{equation}
    e_{i}^{p} = p_{i} - p_{i}^{\prime}(\xi)
    \label{equ:est2}
\end{equation}
, where $p_i$ represented points detected in the current frames and $p_i^\prime(\xi)$ represented points projected from the previous frames into the current frames.

In previous SVO systems \cite{gomez2016pl} \cite{he2018pl}, only the perpendicular re-projection errors of line features were employed in the motion estimation process. This involved calculating the distance from the endpoints of detected line features to the projected infinite line features, expressed as:
\begin{equation}
    e_{j}^{l_{pe}} = \begin{bmatrix}
    d(l_j, p^{\prime}_{j,s}(\xi))\\
    d(l_j, p^{\prime}_{j,e}(\xi))
    \end{bmatrix}
    \label{equ:est3}
\end{equation}
, where $p'_s$ and $p'_e$ represented the endpoints of line features, and $d(.)$ was the distance function from endpoints to lines.

For translational differences between two lines, we employed their midpoints' re-projection residue in the most concise form. Additionally, using midpoints to represent re-projection residuals could facilitate the reuse of point feature code and improve code readability, ultimately saving system running time. Thus, this work introduced re-projection errors parallel to line features for optimization purposes, using the midpoints of line features, i.e.,
\begin{equation}
    e_{k}^{l_{pa}} = p_{i,m} - p_{i,m}^{\prime}(\xi)
\end{equation}
, where $p_{i,m}$ represented the midpoints of the lines detected in the current frames, while $p_{i,m}^\prime(\xi)$ represented the midpoints of the lines projected from the previous frames into the current frames.
By utilizing these two constraint conditions, the re-projection errors between lines could be effectively aligned with their actual pose relationships. This was important as relying on one single constraint for line features might sometimes hinder the ability to ensure consistency between re-projection errors and the true pose relationship between lines.

The optimization problem in equation (\ref{equ:est1}) could be solved iteratively using the Levenberg-Marquardt algorithm. 
    
\subsection{Jacobian Matrix of the Re-Projection Errors of Points and Lines}
\subsubsection{Jacobian of the point re-projection errors}
\label{pointerr}
We used six-dimensional vectors $\xi \in \mathfrak{se}(3)$ to represent the pose transformation matrix $T\in SE(3)$, and the Jacobian of point features was expressed as follows:
\begin{equation}
    J_p = \frac{\partial e_{i}^{p}}{\partial \delta \mathbf{\xi} } = 
    \frac{\partial e_{i}^{p}}{\partial P^\prime }
    \frac{\partial P^\prime}{\partial \delta \mathbf{\xi} }
\end{equation}
, where $P^\prime$ represented the 3D point of matched point feature from the previous frame to the current camera frame. The Jacobian could be divided into two parts using the chain rule. The first part could be expressed by the camera projection principle as follows:
\begin{equation}
    \frac{\partial e_{i}^{p}}{\partial P^\prime} = -\begin{bmatrix}
    \frac{\partial u}{\partial X^\prime} & \frac{\partial u}{\partial Y^\prime} & \frac{\partial u}{\partial Z^\prime}\\
    \frac{\partial v}{\partial X^\prime} & \frac{\partial v}{\partial Y^\prime} & \frac{\partial v}{\partial Z^\prime}\
    \end{bmatrix} = -
    \begin{bmatrix}
    \frac{f_x}{Z^\prime}& 0 & -\frac{f_xX^\prime}{Z^{\prime2}}\\
    0 & \frac{f_y}{Z^\prime} & -\frac{f_yY^\prime}{Z^{\prime2}}
    \end{bmatrix} 
\end{equation}
and the second part could be obtained through the Lie algebra perturbation model:
\begin{equation}
    \frac{\partial P^\prime}{\partial \delta \mathbf{\xi}} = 
    \frac{\partial TP}{\partial \delta \mathbf{\xi} } \Rightarrow 
    \begin{bmatrix}
    I & -P^{\prime\wedge }
    \end{bmatrix}
\end{equation}
, where $[.]^{\wedge}$ denoted skew-symmetric matrix of a vector. 
The Jacobian of point re-projection errors could be rewritten as:
\begin{equation}
\begin{aligned}
    J_p &= \frac{\partial e_{i}^{p}}{\partial \delta \mathbf{\xi} } = \frac{\partial e_{i}^{p}}{\partial P^\prime }
    \frac{\partial P^\prime}{\partial \delta \mathbf{\xi} }
    \\&=-
    \small{
    \setlength{\arraycolsep}{1.2pt}
    \begin{bmatrix}
    \frac{f_x}{Z^{\prime}} & 0 & -\frac{f_xX^{\prime}}{Z^{\prime2}} & -\frac{f_xX^{\prime}Y^{\prime}}{Z^{\prime2}} & f_x + \frac{f_xX^{\prime2}}{Z^{\prime2}} & -\frac{f_xY^{\prime}}{Z^{\prime}}\\
    0 & \frac{f_y}{Z^{\prime}} & -\frac{f_yY^{\prime}}{Z^{\prime2}} &  -f_y - \frac{f_yY^{\prime2}}{Z^{\prime2}} &\frac{f_yX^{\prime}Y^{\prime}}{Z^{\prime2}} &\frac{f_yX^{\prime}}{Z^{\prime}}
    \end{bmatrix}
    \label{equ:3}}
\end{aligned}
\end{equation}
Detailed mathematical derivation could be found in \cite{zuo2017robust} \cite{he2018pl} .
    
\subsubsection{Jacobian of the re-projection errors perpendicular to line features}
The re-projection errors perpendicular to line features were similar to the expression presented in \cite{gomez2016robust} \cite{gomez2019pl}, and were defined in equation (\ref{equ:est3}). Firstly, we converted the 3D line $L_w$ from the world frame to the current camera frame using the following procedure:
\begin{equation}
    L_c = \begin{bmatrix}
    \mathbf{n_c}\\
    \mathbf{d_c}
    \end{bmatrix}
    = T_{cw}L_w = \begin{bmatrix}
    R_{cw} & (t_{cw})^\wedge R_{cw} \\
    0 & R_{cw}
    \end{bmatrix} L_w
\end{equation}
, where $R_{cw}$ and $t_{cw}$ represented the rotation matrix and translation vector, respectively.
Next, the 3D line $L_c$ was projected onto normalized image planes and represented as $l^\prime$, using known intrinsic parameter matrix of cameras, i.e.,
\begin{equation}
    l^{\prime} = \mathcal{K}L_c = \begin{bmatrix}
    f_y & 0 & 0\\
    0 & f_x & 0\\
    -f_yc_x & -f_xc_y &
    f_xf_y\end{bmatrix}\mathbf{n_c} = \begin{bmatrix}
    l_1\\
    l_2\\
    l_3
    \end{bmatrix}
\end{equation}
, where $\mathcal{K}$ represented the projection matrix of the line. Since we projected the line features onto the image plane as an infinite line, only the normal component $\mathbf{n_c}$ in the $Pl\ddot{u}cker$ coordinates $L_c$ provided meaningful information during projection.
As mentioned in Section \ref{4A}, the re-projection errors perpendicular to line features could be expressed as follows:
\begin{equation}
    e_{j}^{l_{pe}} = \begin{bmatrix}
    d_s\\
    d_e
    \end{bmatrix} = \begin{bmatrix}
    \frac{p^{T}_sl^{\prime}}{\sqrt{l^2_1+l^2_2} }\\
    \frac{p^{T}_el^{\prime}}{\sqrt{l^2_1+l^2_2} }
    \end{bmatrix}
\end{equation}
We assumed $l = \sqrt{l^2_1+l^2_2}$ and $d = \frac{\mathbf{p^{T}}l^{\prime}}{l}$, and the Jacobian of perpendicular line re-projection errors could be expressed as:
\begin{equation}
    \begin{aligned}
    \frac{\partial d}{\partial \delta \xi } = \frac{\partial \frac{\mathbf{p^{T}}l^{\prime}}{l}}{\partial \delta \xi}
     & = \frac{\partial (u\,l_1 + v\,l_2 + l_3)}{\partial \delta \xi}\frac{1}{l}\\
    & = \begin{bmatrix}l1 & l2\end{bmatrix}
    \frac{\partial \begin{bmatrix}u \\ v\end{bmatrix}}{\partial \delta \xi}
    \frac{1}{l}
    \end{aligned}   
\end{equation}
Equation below could be obtained using the chain rule:
\begin{equation}
    \frac{\partial \begin{bmatrix}u \\ v\end{bmatrix}}{\partial \delta \xi} = \frac{\partial \begin{bmatrix}u \\ v\end{bmatrix}}{\partial P^{\prime}}\frac{\partial P^{\prime}}{\partial \delta \xi}
\end{equation}
    
Referring to equation (\ref{equ:3}), the Jacobian of perpendicular line re-projection errors could be expressed as follows:
\begin{equation}
\begin{aligned}
    \frac{\partial d}{\partial \delta \xi } &=-
    \begin{bmatrix}l1 & l2\end{bmatrix} \\&
    \small{
    \setlength{\arraycolsep}{1.2pt}
    \begin{bmatrix}
    \frac{f_x}{Z^{\prime}} & 0 & -\frac{f_xX^{\prime}}{Z^{\prime2}} & -\frac{f_xX^{\prime}Y^{\prime}}{Z^{\prime2}} & f_x + \frac{f_xX^{\prime2}}{Z^{\prime2}} & -\frac{f_xY^{\prime}}{Z^{\prime}} \\
    0 & \frac{f_y}{Z^{\prime}} & -\frac{f_yY^{\prime}}{Z^{\prime2}} &  -f_y - \frac{f_yY^{\prime2}}{Z^{\prime2}} &\frac{f_yX^{\prime}Y^{\prime}}{Z^{\prime2}} &\frac{f_yX^{\prime}}{Z^{\prime}}
    \end{bmatrix}}
    \frac{1}{l}&
\end{aligned}
\end{equation}
    
\begin{equation}
    J_{l_{pe}} = d_s  \frac{\partial d_s}{\partial \delta \xi } 
    + d_e  \frac{\partial d_e}{\partial \delta \xi } 
\end{equation}

In comparison to methods of directly deriving the Lie algebra of the transformation, as presented in \cite{zuo2017robust} \cite{he2018pl}, obtaining the Jacobian of re-projection errors perpendicular to line features was more efficient and convenient over ones using derivation results from line endpoints.
\subsubsection{Jacobian of the re-projection errors parallel to line features}
Similar to previous section, we defined the re-projection errors of matched and projected line midpoints as the cost function for the re-projection errors parallel to line features, i.e.,
\begin{equation}
    \frac{\partial e_{k}^{l_{pa}}}{\partial \delta \mathbf{\xi} } = 
    \frac{\partial e_{k}^{l_{pa}}}{\partial P_{m}^{\prime} }
    \frac{\partial P_{m}^{\prime}}{\partial \delta \mathbf{\xi} }
\end{equation}
, where $P_m^\prime$ represented the 3D midpoint of the matched line feature from the previous frame to the current frame.
Referring to (\ref{equ:3}), we could obtain Jacobian of the re-projection errors parallel to line features as follows:
\begin{equation}
\begin{aligned}
    J_{l_{pa}} &= \frac{\partial e_{k}^{l_{pa}}}{\partial \delta \mathbf{\xi} } 
    \\& = -
    \small{
    \setlength{\arraycolsep}{1.2pt}
    \begin{bmatrix}\frac{f_x}{Z_m^{\prime}} & 0 & -\frac{f_xX_m^{\prime}}{Z_m^{\prime2}} & -\frac{f_xX_m^{\prime}Y_m^{\prime}}{Z_m^{\prime2}} & f_x + \frac{f_xX_m^{\prime2}}{Z_m^{\prime2}} & -\frac{f_xY_m^{\prime}}{Z_m^{\prime}}\\
    0 & \frac{f_y}{Z_m^{\prime}} & -\frac{f_yY_m^{\prime}}{Z_m^{\prime2}} &  -f_y - \frac{f_yY_m^{\prime2}}{Z_m^{\prime2}} &\frac{f_yX_m^{\prime}Y_m^{\prime}}{Z_m^{\prime2}} &\frac{f_yX_m^{\prime}}{Z_m^{\prime}}
    \end{bmatrix}}
\end{aligned}
\end{equation}

Thus far, we have obtained the Jacobians for the re-projection errors of all three types of features. The pose transformation between adjacent frames could be achieved by solving the non-linear least-square equation (\ref{equ:est1}) using the Levenberg-Marquardt algorithm.

\section{Experimental Results and Analysis}
For fair comparison, we used only the front end of PL\_SLAM, ORB\_SLAM2, and ORB\_Line SLAM to form VO systems to avoid possible influences caused by loop closure detection in these systems. All experiments were conducted on an Intel Core i5-4210U CPU @ 1.70GHz × 4 and 16GB RAM without GPU acceleration. 

\subsection{Performance on KITTI}
We tested the DynPL-SVO on the KITTI dataset, which provided ground truth trajectories based on a 64-channel Velodyne LiDAR sensor and GPS localization. Additionally, the presence of dynamic scenes containing moving objects such as cars and pedestrians in some sequences had a significant impact on the performance of VO/vSLAM systems.

We presented the absolute pose error (APE) in Table \ref{table1:exp} along with three other benchmark systems, wherein we listed the root-mean-square error (RMSE) of absolute translation and rotation errors for all methods. It showed that the DynPL-SVO outperformed other methods in 9 sequences, demonstrating its superior accuracy in motion estimation across most sequences, especially those that were highly dynamic, such as sequences 01, 05, and 09. The translation RMSE drifts of the DynPL-SVO were improved by 13.8\%, 30.0\%, and 24.8\% averagely compared to those of PL\_SLAM front end, ORB\_SLAM2 front end, and ORB\_Line SLAM front end, respectively, indicating its superior efficiency.

\begin{figure*}[!t]  
    \centering    
    \hfill
    \subfloat[Top view of sequence 00] 
    {
        \begin{minipage}[!t]{0.32\textwidth}
            \centering          
            \includegraphics[width=1\textwidth]{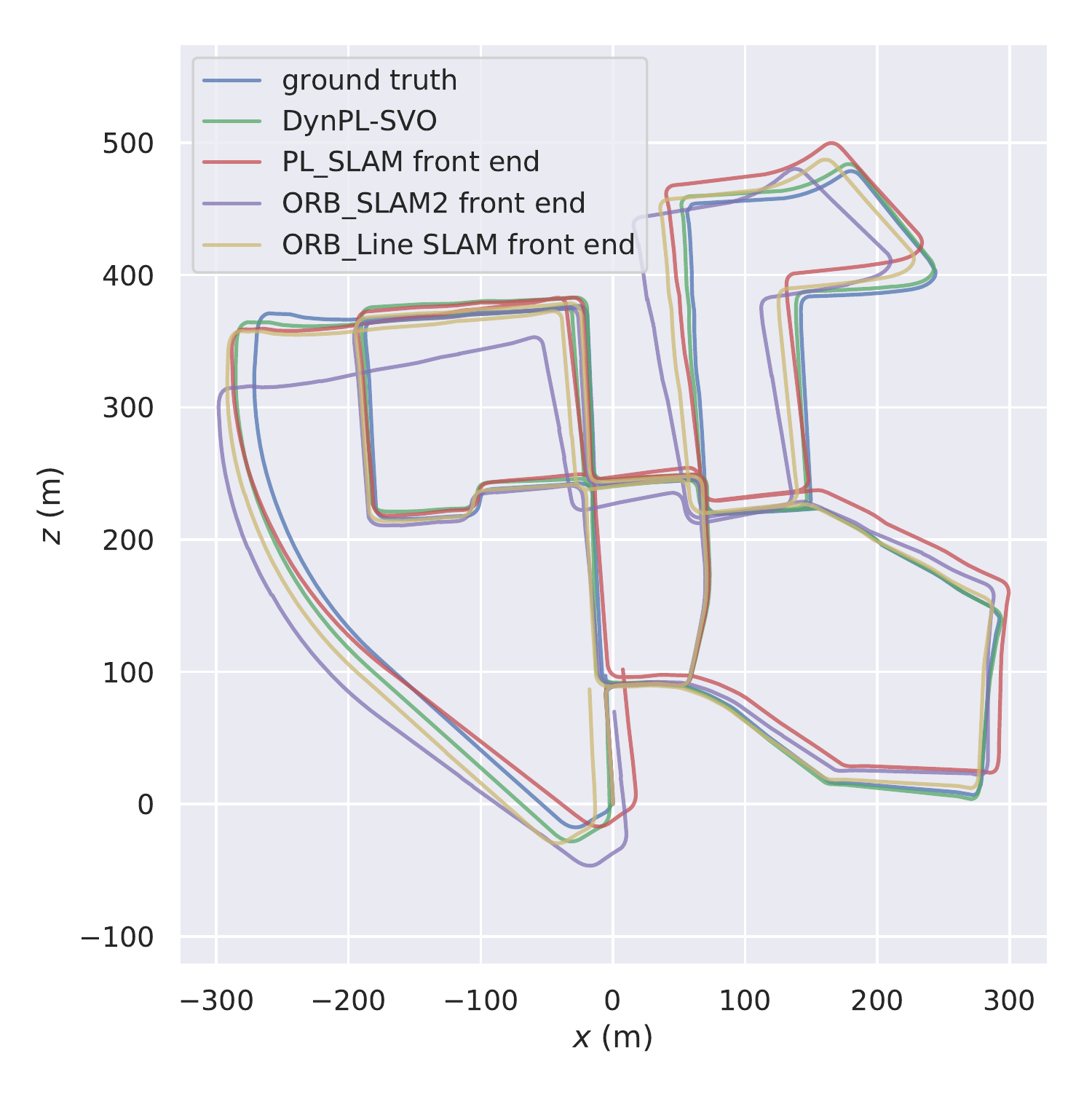}   
        \end{minipage}%
    }\hfill
    \subfloat[Top view of sequence 05] 
    {
        \begin{minipage}[!t]{0.32\textwidth}
            \centering      
            \includegraphics[width=1\textwidth]{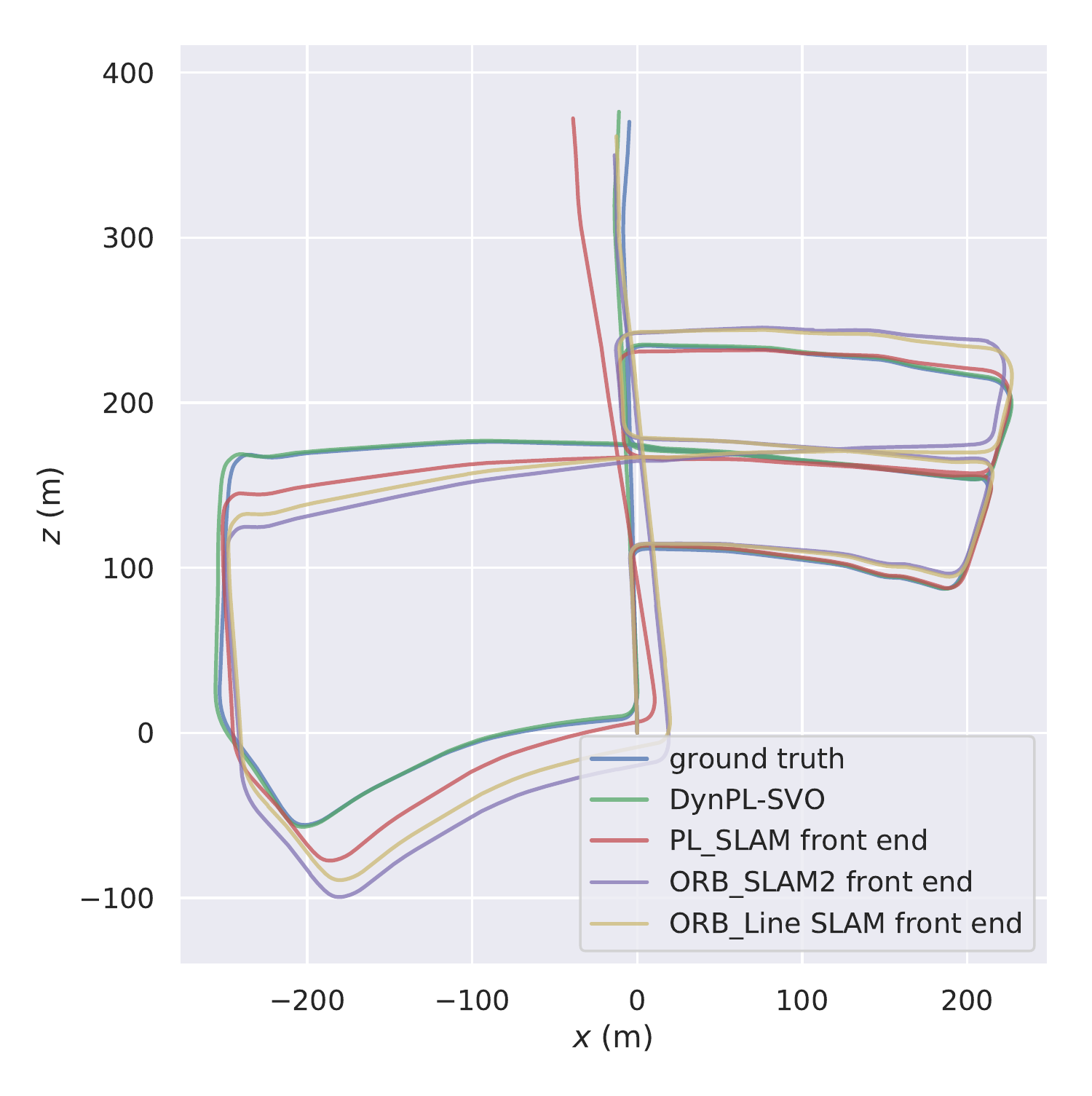}   
        \end{minipage}
    }\hfill
    \subfloat[Top view of sequence 08] 
    {
        \begin{minipage}[!t]{0.32\textwidth}
            \centering      
            \includegraphics[width=1\textwidth]{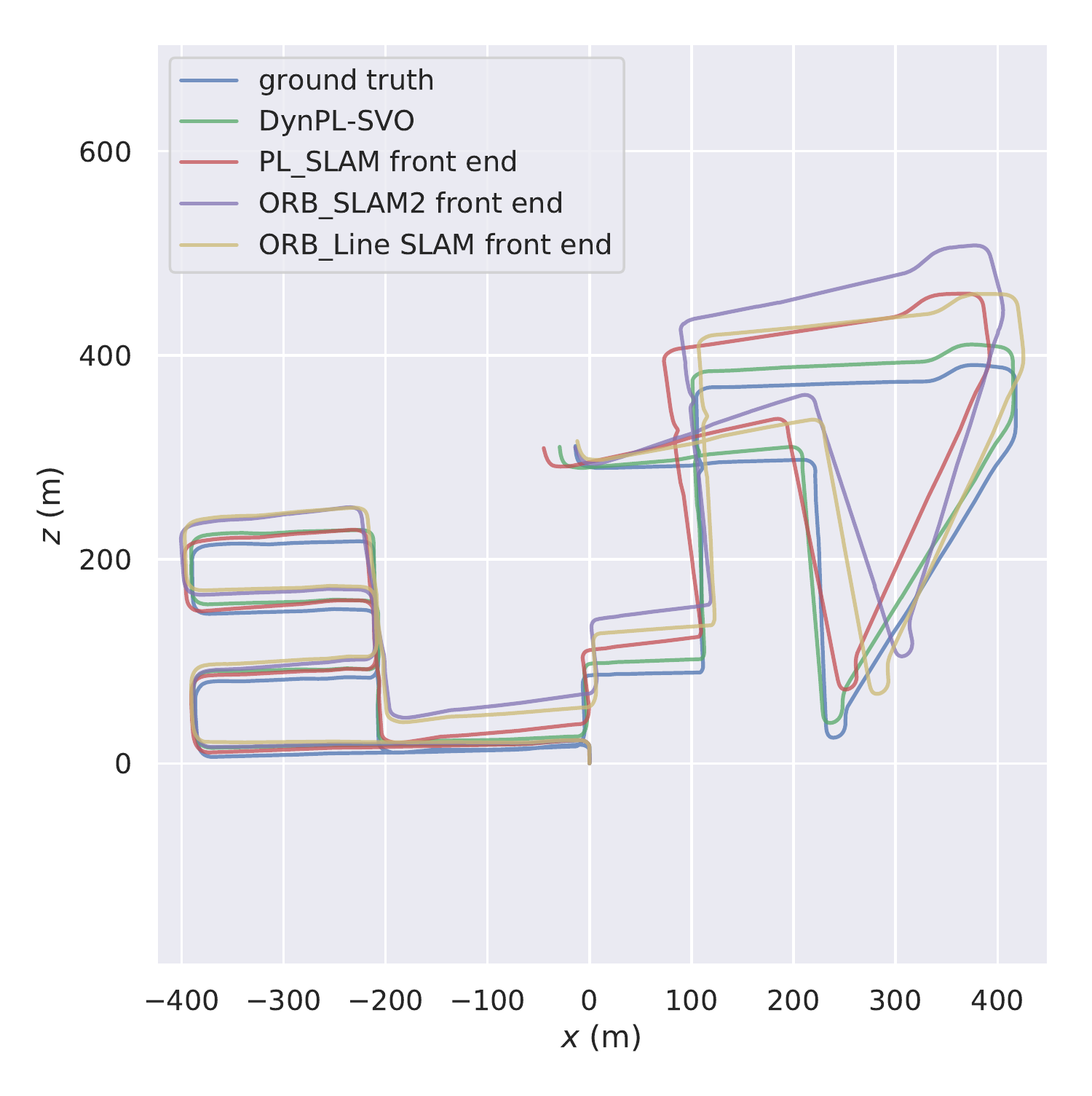}   
        \end{minipage}
    }\hfill
    \caption{Reconstruction of the path from DynPL-SVO, PL\_SLAM front end, ORB\_SLAM2 front end, and ORB\_Line SLAM front end. The blue lines represented the ground truth trajectory, the green lines corresponded to the estimation of DynPL-SVO, while the red lines represented the estimation of PL\_SLAM front end. The paths provided by ORB\_SLAM2 and ORB\_Line SLAM were plotted with purple and brown lines, respectively.
    } 
    \label{fig1:exp}  
\end{figure*}

Table \ref{table1-2:exp} presented the relative pose error (RPE) comparisons, where the DynPL-SVO achieved better translation and rotation accuracy over other comparative systems in most scenes, with an average improvement of 14.8\% and 2.1\%, respectively, further confirming its superior performance. Figure \ref{fig1:exp} depicted the reconstructed paths of the four SVO systems on several sequences of the KITTI dataset, wherein the ones from benchmark methods had larger deviations over the DynPL-SVO in all three sequences, particularly around corners with large viewpoint changes. This further confirmed that the DynPL-SVO outperformed other SVO systems, specifically in dealing with dynamic environments.

\begin{table*}[!t]
\centering
 \caption{Mean Absolute RMSE in the KITTI Dataset, with the dash indicating failed experiment.}
 \begin{threeparttable}
 \begin{tabular}{c | c  c | c  c | c  c | c  c} 
 \hline
 \multicolumn{1}{c|}{\multirow{2}*{Seq.}}&\multicolumn{2}{c|}{ DynPL-SVO}&\multicolumn{2}{c|}{PL\_SLAM front end}&\multicolumn{2}{c|}{ORB\_SLAM2 front end}&\multicolumn{2}{c}{ORB\_Line SLAM front end}\\
 ~ & t$_{m}$ & R$_{deg}$ & t$_{m}$ & R$_{deg}$ & t$_{m}$ & R$_{deg}$ & t$_{m}$ & R$_{deg}$\\ [0.5ex] 
 \hline\hline
 00 & \textbf{6.691} & 1.788 & 7.426 & 2.105 & 14.076 & 3.461 & 7.551 & \textbf{1.519}\\
 01 & \textbf{172.502} & \textbf{8.910} & 371.245 & 12.212 & - & - & - & -\\
 02 & 21.653 & 4.423 & \textbf{8.167} & \textbf{1.505} & 14.187 & 2.615 & 11.276 & 2.550\\
 03 & 6.077 & 4.308 & 6.030 & 3.310 & \textbf{2.317} & 1.511 & 3.018 & \textbf{1.466}\\
 04 & \textbf{2.100} & \textbf{29.944} & 2.216 & 34.067 & 2.655 & 49.025 & 2.550 & 38.861\\
 05 & \textbf{4.097} & \textbf{1.598} & 6.506 & 2.695 & 11.755 & 4.217 & 8.750 & 3.641\\
 06 & \textbf{4.113} & 2.927 & 5.564 & 6.305 & 4.219 & \textbf{1.518} & 4.120 & 2.364\\
 07 & 5.216 & \textbf{1.957} & \textbf{3.028} & 2.165 & 14.155 & 5.698 & 15.512 & 6.957\\
 08 & \textbf{7.202} & \textbf{3.019} & 10.054 & 3.254 & 24.796 & 6.134 & 17.134 & 3.302\\
 09 & \textbf{4.729} & \textbf{1.065} & 12.205 & 2.454 & 18.387 & 3.718 & 24.154 & 4.288\\
 10 & \textbf{2.064} & 1.831 & 2.649 & \textbf{1.155} & 3.823 & 2.081 & 4.992 & 1.823\\
 \hline
\end{tabular}
\label{table1:exp}
    \begin{tablenotes} 
		\item t$_{m}$:average translational RMSE drift(meter).
        \item R$_{deg}$:average rotational RMSE drift ($^\circ$).
    \end{tablenotes} 
\end{threeparttable}
\end{table*}

\begin{table*}[!t]
\centering
 \caption{Mean Relative pose errors on the KITTI Dataset, with the dash indicating failed experiment.}
  \begin{threeparttable}
 \begin{tabular}{c | c  c | c  c | c  c| c   c} 
 \hline
 \multicolumn{1}{c|}{\multirow{2}*{Seq.}}&\multicolumn{2}{c|}{ DynPL-SVO}&\multicolumn{2}{c|}{PL\_SLAM front end}&\multicolumn{2}{c|}{ORB\_SLAM2 front end}&\multicolumn{2}{c}{ORB\_Line SLAM front end}\\
 ~ & t$_{\%}$ & R$_{deg/100m}$ & t$_{\%}$ & R$_{deg/100m}$ & t$_{\%}$ & R$_{deg/100m}$ & t$_{\%}$ & R$_{deg/100m}$\\ [0.5ex] 
 \hline\hline
00 & 1.569 & 0.443 & 1.607 & 0.405 & 1.424 & 0.590 & \textbf{1.137} & \textbf{0.366}\\
01 & \textbf{21.339} & \textbf{1.402} & 43.723 & 1.939 & - & - & - & -\\
02 & 1.733 & 0.517 & 1.738 & \textbf{0.344} & 1.621 & 0.508 & \textbf{1.581} & 0.433\\
03 & 3.604 & 1.637 & 3.467 & 1.260 & \textbf{1.964} & 0.712 & 2.140 & \textbf{0.639}\\
04 & \textbf{1.907} & 0.424 & 2.023 & \textbf{0.282} & 2.447 & 0.498 & 3.398 & 0.511\\
05 & \textbf{1.007} & \textbf{0.375} & 1.412 & 0.493 & 2.649 & 0.754 & 2.405 & 0.609\\
06 & 1.898 & 0.582 & 2.372 & \textbf{0.506} & 1.974 & 0.524 & \textbf{1.761} & 0.567\\
07 & 2.051 & \textbf{0.879} & \textbf{1.725} & 1.047 & 4.658 & 2.138 & 5.067 & 2.542\\
08 & \textbf{1.279} & \textbf{0.453} & 1.670 & 0.468 & 3.121 & 0.959 & 2.642 & 0.608\\
09 & \textbf{1.486} & \textbf{0.353} & 2.129 & 0.465 & 3.422	& 1.044 & 4.388 & 0.933\\
10 & 1.204 & 0.572 & \textbf{1.032} & \textbf{0.337} & 1.693 & 0.716 & 1.903 & 0.566\\
\hline
\end{tabular}
\label{table1-2:exp}
    \begin{tablenotes} 
		\item t$_{\%}$:average translational RMSE drift(\%).
        \item R$_{deg/100m}$:average rotational RMSE drift ($^\circ$/100m).
    \end{tablenotes} 
\end{threeparttable}
\end{table*}

In addition, we conducted an ablation study of the DynPL-SVO. Table \ref{table2:exp} showed that the DynPL-SVO with only re-projection errors parallel to line features achieved better accuracy in 10 sequences compared to the system without line re-projection, and improved accuracy by 30.9\% as compared to those using only the re-projection errors perpendicular to line features in terms of RPE on the KITTI dataset, illustrating importance of considering them in SVO systems.
\subsection{Performance on EuRoC MAV}
We compared the  DynPL-SVO and PL\_SLAM front end and performed an ablation study on the EuRoC MAV dataset \cite{liu2021plc} to validate the effects of various line re-projection errors on estimation accuracy. In Table \ref{table3:exp}, we showed that re-projection errors parallel to line features were more important over re-projection errors perpendicular to line features in most sequences, with an 8.6\% improvement overall. The DynPL-SVO outperformed PL\_SLAM front end in several sequences, including MH\_01\_easy and MH\_04\_different, due to the presence of many short but complete line features in these sequences, wherein the introduction of the re-projection errors parallel to line features in the DynPL-SVO helped improve its performance. It is worth noting that the introduction of re-projection errors perpendicular to line features had a negative impact on three sequences. This could be attributed to irregular drone motions and numerous short line features in the scenes, leading to misalignment and mismatches.

\begin{table*}[!t]
\centering
 \caption{Mean Absolute and Relative RMSE errors of the DynPL-SVO on the KITTI Dataset.}
 \begin{tabular}{c | c  c | c  c | c  c} 
 \hline
    \multicolumn{1}{c|}{\multirow{2}*{Seq.}}&\multicolumn{2}{c|}{ wo/line error}&\multicolumn{2}{c|}{\tabincell{c}{w/re-projection errors \\ perpendicular to line features}}&\multicolumn{2}{c}{\tabincell{c}{w/re-projection errors \\ parallel to line features}}\\
    ~ & APE & RPE & APE & RPE & APE & RPE\\
 \hline\hline
 00 & 7.3728 & 0.0323 & 11.9514 & 0.0391 & \textbf{6.6818} & \textbf{0.0322}\\
 01 & \textbf{52.5071} & \textbf{0.7188} & 166.6740 & 1.0059 & 89.8455 & 0.7788 \\
 02 & 20.6018 & 0.0348 & 23.6016 & 0.0559 & \textbf{19.5024} & \textbf{0.0346} \\
 03 & 6.2301 & \textbf{0.0317} & \textbf{6.0842} & 0.0322 & 6.2207 & \textbf{0.0317} \\
 04 & 2.7768 & 0.0374 & \textbf{2.2843} & 0.0554 & 2.7410 & \textbf{0.0365} \\
 05 & 3.9010 & 0.0182 & 4.3099 & 0.0196 & \textbf{3.8735} & \textbf{0.0181} \\
 06 & 5.0703 & 0.0326 & \textbf{4.6390} & 0.0448 & 4.9705 & \textbf{0.0322} \\
 07 & \textbf{1.5365} & 0.0178 & 5.4844 & 0.0634 & 1.8679 & \textbf{0.0174} \\
 08 & 5.4780 & 0.0390 & 7.3161 & 0.0441 & \textbf{4.8398} & \textbf{0.0389} \\
 09 & 4.5855 & 0.0248 & \textbf{4.4380} & 0.0674 & 4.5637 & \textbf{0.0245} \\
 10 & 2.0184 & 0.0198 & 2.2578 & 0.0342 & \textbf{2.0173} & \textbf{0.0197} \\
 \hline
\end{tabular}
\label{table2:exp}
\end{table*}

\begin{table*}[!t]
\centering
 \caption{Mean Relative RMSE of the DynPL-SVO on the EuRoC MAV dataset.}
 \begin{threeparttable}
 \begin{tabular}{c | c | c | c} 
 \hline
 Seq. & \tabincell{c}{PL\_SLAM \\ front end} & \tabincell{c}{DynPL-SVO w/only re-projection errors \\ perpendicular to line features} & \tabincell{c}{DynPL-SVO w/only re-projection errors \\ parallel to line features}  \\[0.5ex] 
 \hline\hline
 MH\_01\_easy & 0.033349 & \textbf{0.033294} & 0.033358\\
 MH\_02\_easy & 0.032896 & 0.032501 & \textbf{0.032016}\\
 MH\_03\_med & 0.073692 & 0.071908 & \textbf{0.071196}\\
 MH\_04\_dif & 0.103936 & \textbf{0.103346} & 0.103473\\
 MH\_05\_dif & 0.095603 & 0.094640 & \textbf{0.094608}\\
 V1\_01\_easy & \textbf{0.048642} & 0.049011 & 0.049427\\
 V1\_02\_med & 0.102017 & 0.102260 & \textbf{0.101900}\\
 V1\_03\_dif & 0.098576 & 0.101670 & \textbf{0.097804}\\
 V2\_01\_easy & 0.037155 & 0.032655 & \textbf{0.032629}\\
 V2\_02\_med & 0.074001 & 0.073592 & \textbf{0.071348}\\
 \hline
\end{tabular}
\label{table3:exp}
    \begin{tablenotes} 
		\item The sequence V2\_03\_difficult contained an unequal number of left and right images, rendering it unsuitable for evaluating stereo systems, and therefore, it was not included in the table presented above.
     \end{tablenotes} 
\end{threeparttable}
\end{table*}

\subsection{Evaluating the capability of dynamic grids in dealing with dynamic scenes}
\begin{figure*}[!t]  
    \centering    
    \subfloat[] 
    {
        \begin{minipage}[!t]{0.45\textwidth}
            \centering          
            \includegraphics[width=\textwidth]{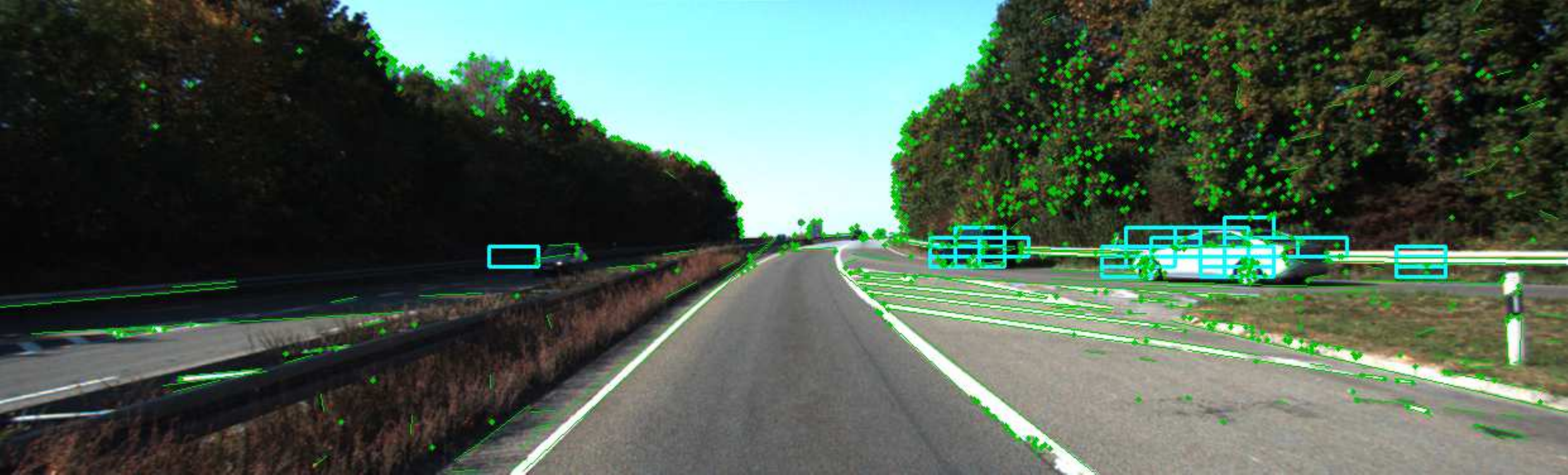}   
        \end{minipage}%
    }
    \subfloat[] 
    {
        \begin{minipage}[!t]{0.45\textwidth}
            \centering      
            \includegraphics[width=1\textwidth]{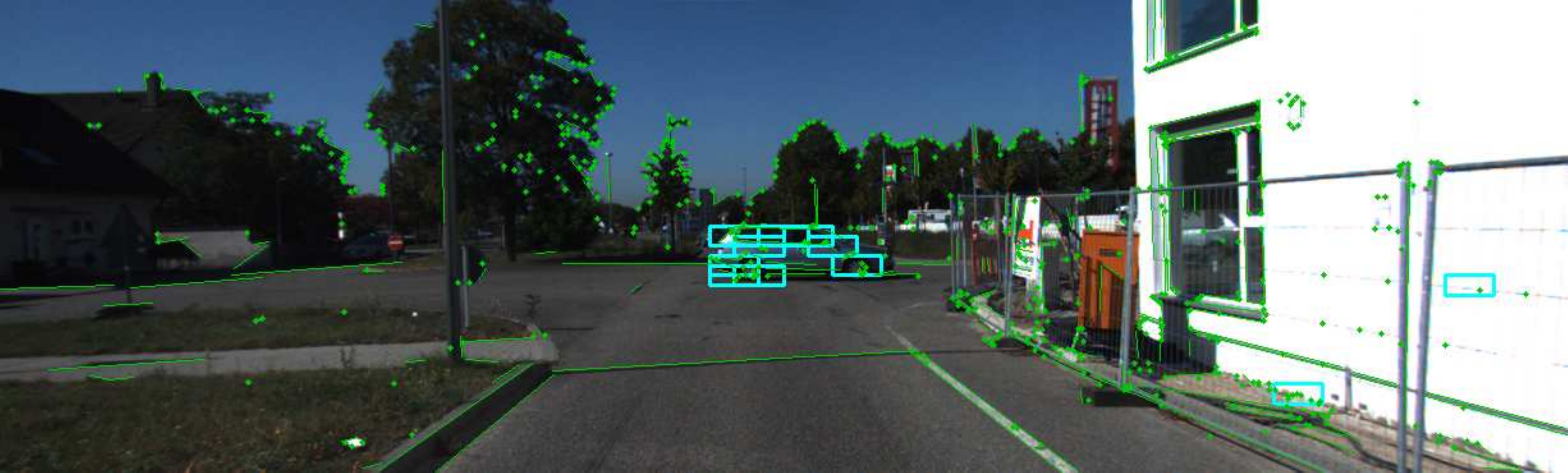}   
        \end{minipage}
    }
    
    \subfloat[] 
    {
        \begin{minipage}[!t]{0.45\textwidth}
            \centering      
            \includegraphics[width=1\textwidth]{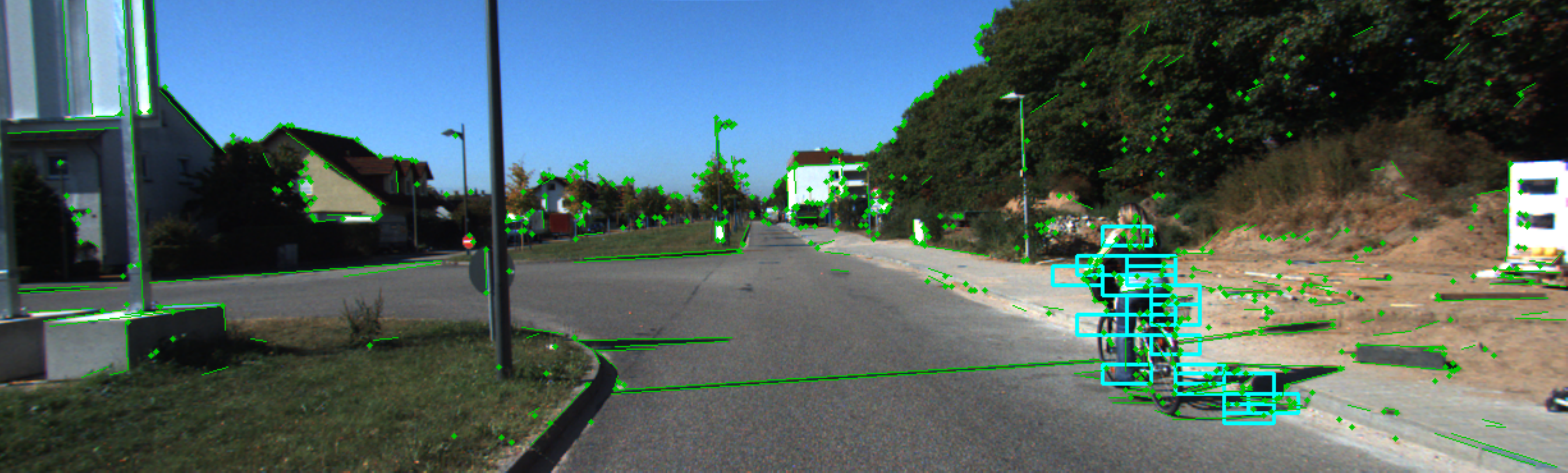}   
        \end{minipage}
    }
    \subfloat[] 
    {
        \begin{minipage}[!t]{0.45\textwidth}
            \centering      
            \includegraphics[width=1\textwidth]{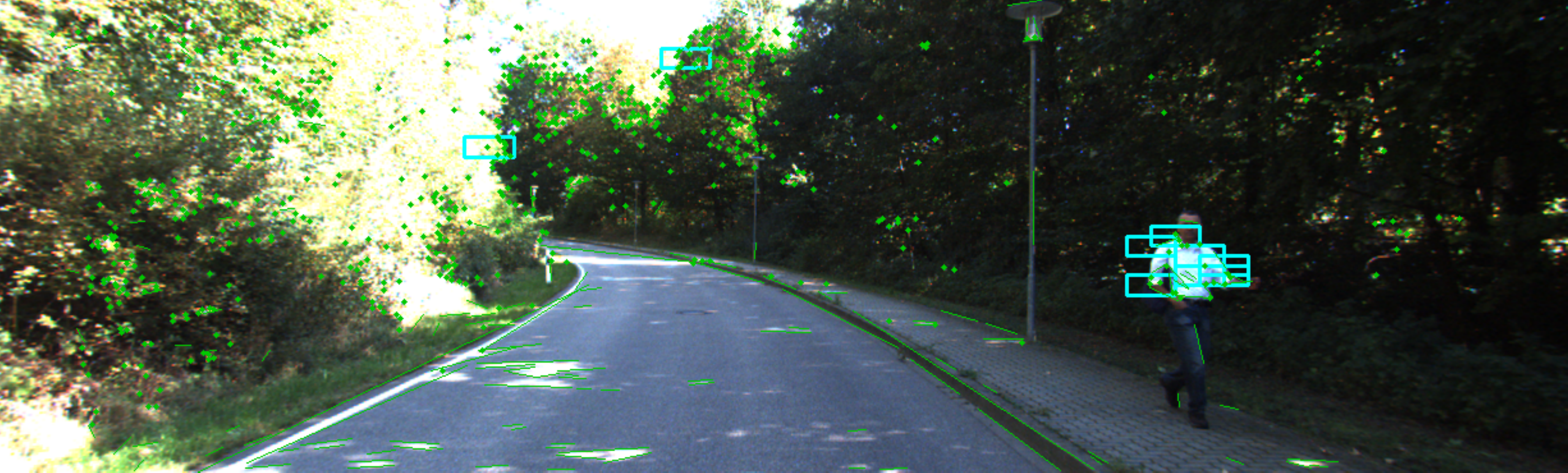}   
        \end{minipage}
    }
    \caption{Dynamic scenes in the \textit{KITTI} dataset and the impact of \textit{dynamic grid}s on dynamic regions. In each subfigure, cyan boxes represented the \textit{dynamic grid}s. Figure \ref{fig5:exp}(a) and \ref{fig5:exp}(b) showed moving cars in \textit{KITTI-01} and \textit{KITTI-06}, respectively. Figure \ref{fig5:exp}(c) showcased a person riding bike in \textit{KITTI-06}, while Figure \ref{fig5:exp}(d) depicted a pedestrian in \textit{KITTI-09}.} 
    \label{fig5:exp}  
\end{figure*}

We conducted a comparative analysis in order to evaluate the effectiveness of the \textit{dynamic grid}s. As shown in Figure \ref{fig5:exp}(a) and \ref{fig5:exp}(b), \textit{dynamic grid}s could identify vehicles traveling at high speeds in different directions in the KITTI dataset, with or without rich structural information. Additionally, for slow-moving objects such as cyclists and pedestrians, \textit{dynamic grid}s could eliminate dynamic features, as demonstrated in Figure \ref{fig5:exp}(c) and \ref{fig5:exp}(d), indicating that the \textit{dynamic grid}s could accurately identify dynamic regions, reducing the influence of dynamic features on accuracy, and making the DynPL-SVO more robust when operating in dynamic scenes.

Table \ref{table4:exp} presented a quantitative analysis of the \textit{dynamic grid} approaches in the KITTI dataset. Results illustrated that the DynPL-SVO with the \textit{dynamic grid} method provided more accurate estimation in 8 sequences, improving APE and RPE by about 13.6\% and 2.3\%, respectively, compared to those without the \textit{dynamic grid}s. It was noteworthy that the \textit{dynamic grid} method worked well, particularly in sequences with highly dynamic scenes such as KITTI-01, 05, and 09, wherein SVO accuracy was improved by over 30\%.

\begin{table}[!t]
\centering
 \caption{Mean Absolute and Relative RMSE errors of the DynPL-SVO in KITTI Dataset.}
 
 \begin{tabular}{c | c  c | c  c} 
 \hline
    \multicolumn{1}{c|}{\multirow{2}*{Seq.}}&\multicolumn{2}{c|}{\tabincell{c}{DynPL-SVO \\ w/\textit{dynamic grid}}}&\multicolumn{2}{c}{\tabincell{c}{DynPL-SVO \\ wo/\textit{dynamic grid}}}\\
    ~ & APE & RPE & APE & RPE\\
 \hline\hline
 00 & \textbf{11.221715} & 0.040752 & 13.236462 & \textbf{0.038351}\\
 01 & \textbf{182.531449} & \textbf{1.019784} & 319.707988 & 1.306922\\
 02 & 21.654051 & 0.064572 & \textbf{10.960957} & \textbf{0.055898}\\
 03 & 6.094158 & 0.032278 & \textbf{5.946504} & \textbf{0.031802}\\
 04 & 2.207623 & 0.054315 & \textbf{2.015523} & \textbf{0.052209}\\
 05 & \textbf{4.403628} & 0.019534 & 7.060361 & \textbf{0.019145}\\
 06 & \textbf{4.428359} & \textbf{0.044869} & 4.440736 & 0.056673\\
 07 & 5.216922 & \textbf{0.062928} & \textbf{4.725139} & 0.064078\\
 08 & \textbf{7.443639} & 0.043877 & 12.509229 & \textbf{0.042731}\\
 09 & \textbf{4.823906} & \textbf{0.066494} & 13.612393 & 0.070549\\
 10 & \textbf{2.282632} & \textbf{0.032286} & 2.786268 & 0.033558\\
 \hline
\end{tabular}
\label{table4:exp}
\end{table}

\section{Conclusions}
In this paper, we proposed a robust SVO method, i.e., DynPL-SVO, that utilized both point and line features to improve motion estimation accuracy in dynamic scenes. The method introduced the re-projection errors parallel to line features into cost functions to make full use of the structural information of line features. The \textit{dynamic grid} method was also introduced to address the reduction of robustness and accuracy of SVO systems caused by moving objects, wherein dynamic regions could be efficiently marked and point features on dynamic objects could be removed without using depth information and other sensors.
The performance of the DynPL-SVO was compared with three SOTA SVO systems on two datasets. Comprehensive experimental results showed that the DynPL-SVO achieved more robust and accurate results in most scenes, particularly on highly dynamic scenes.

Future research will focus on introducing features with greater geometric information such as planes and cubes into VO systems to further improve estimation accuracy of SVO systems.

\bibliographystyle{ieeetr}
\bibliography{references}

\end{document}